\documentclass[10pt,twocolumn,letterpaper]{article}

\usepackage{iccv}
\usepackage{times}
\usepackage{epsfig}
\usepackage{graphicx}
\usepackage{amsmath}
\usepackage{amssymb}

\usepackage[pagebackref=true,breaklinks=true,letterpaper=true,colorlinks,bookmarks=false]{hyperref}
\usepackage{cite}
\usepackage{bbding}
\usepackage{multirow}
\usepackage{booktabs}
\usepackage{makecell}

\iccvfinalcopy 

\def\iccvPaperID{****} 

\ificcvfinal\fi

\begin{document}


\title{ELITE: Encoding Visual Concepts into Textual Embeddings \\ for Customized Text-to-Image Generation}

\author{ Yuxiang Wei$^{1, 2}$ \quad
Yabo Zhang$^{1}$ \quad
Zhilong Ji$^{3}$ \quad
Jinfeng Bai$^{3}$ \quad
Lei Zhang$^{2}$ \quad
Wangmeng Zuo$^{1, 4}$\textsuperscript{(\Envelope)} \\
\small
$^1$Harbin Institute of Technology \quad $^2$ The Hong Kong Polytechnic University  \quad $^3$Tomorrow Advancing Life \quad $^4$ Peng Cheng Lab  \\
}

\maketitle
\ificcvfinal\thispagestyle{empty}\fi

\begin{abstract}
In addition to the unprecedented ability in imaginary creation, large text-to-image models are expected to take customized concepts in image generation.
Existing works generally learn such concepts in an optimization-based manner, yet bringing excessive computation or memory burden.
%
In this paper, we instead propose a learning-based encoder, which consists of a global and a local mapping networks for fast and accurate customized text-to-image generation.
In specific, the global mapping network projects the hierarchical features of a given image into multiple ``new'' words in the textual word embedding space, i.e., one primary word for well-editable concept and other auxiliary words to exclude irrelevant disturbances (e.g., background).
In the meantime, a local mapping network injects the encoded patch features into cross attention layers to provide omitted details, without sacrificing the editability of primary concepts.
We compare our method with existing optimization-based approaches on a variety of user-defined concepts, and demonstrate that our method enables high-fidelity inversion and more robust editability with a significantly faster encoding process.
Our code is publicly available at \url{https://github.com/csyxwei/ELITE}.


\end{abstract}

\section{Introduction}

Recently, large-scale diffusion models~\cite{saharia2022photorealistic, ramesh2022hierarchical, balaji2022ediffi, rombach2022high} have demonstrated impressive superiority in text-to-image generation.
By training with billions of image-text pairs, large text-to-image diffusion models have exhibited excellent semantic understanding ability, and  generate diverse and photo-realistic images being accordant to the given text prompts.
Owning to their unprecedentedly creative capabilities, these models have been applied to various tasks, such as image editing~\cite{hertz2022prompt,mokady2022null}, data augmentation~\cite{ni2022imaginarynet}, and even artistic creation~\cite{pan2022synthesizing}.

\begin{figure}[t]
\centering
\includegraphics[width=1\linewidth]{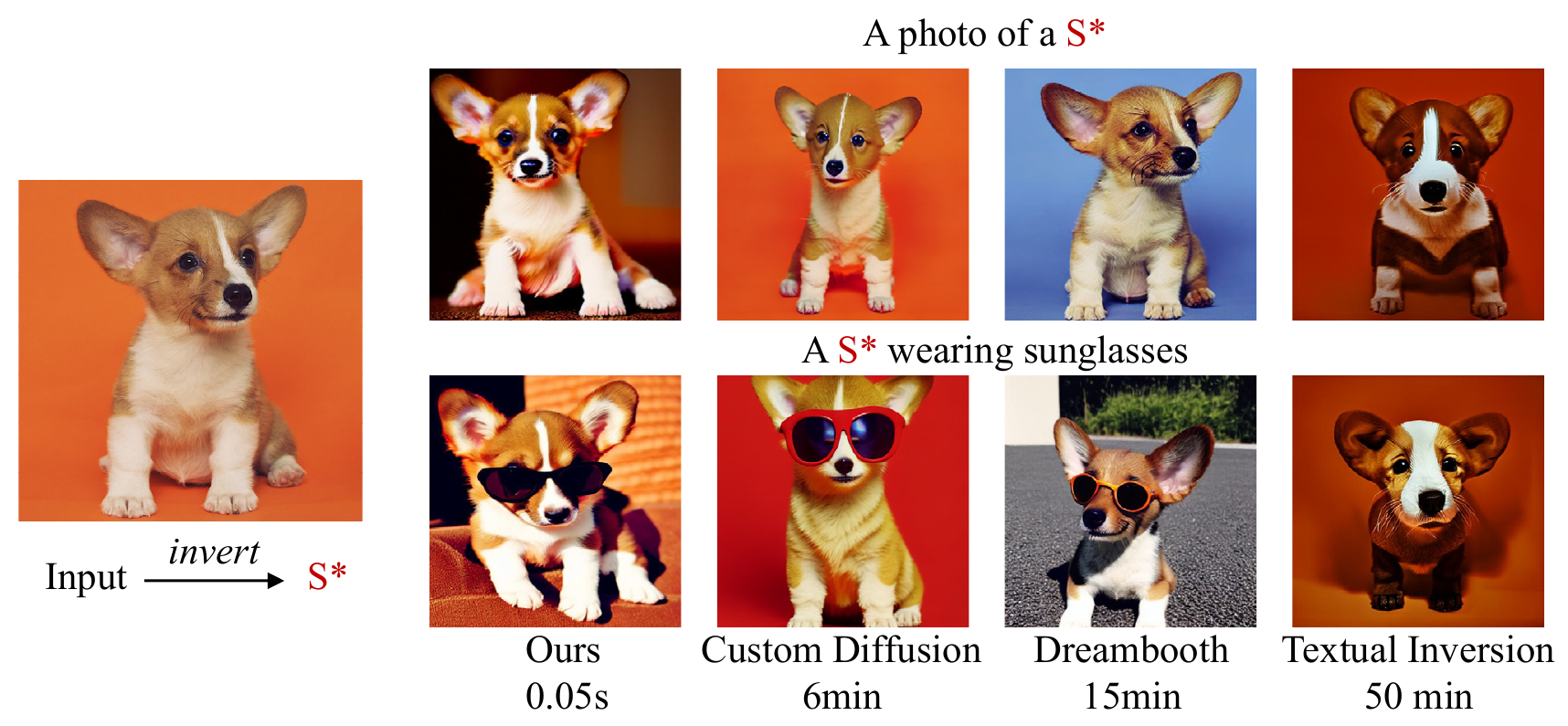}
\caption{Given an input image, customized text-to-image generation learns a pseudo-word (S*) in word embedding space to represent the target concept. With S*, one can flexibly synthesize or edit the concept with text prompts. The running time to learn a new concept is also listed, and our method learns the new concept much faster than others.}
\label{fig:comparison}
\end{figure}

However, despite diverse and general generation, users may expect to create imaginary instantiates with undescribable personalized concepts~\cite{kumari2022multi}, \eg, ``corgi'' in Fig.~\ref{fig:comparison}.
To this end, many recent studies have been conducted for customized text-to-image generation~\cite{gal2022image, ruiz2022dreambooth, kumari2022multi, cohen2022my}, which aims to learn a specific concept from a small set of user-provided images (\eg, 3$\sim$5 images).
Then, users can flexibly compose the learned concepts into new scenes, \eg, \texttt{A S* wearing sunglasses} in Fig.~\ref{fig:comparison}.
Given a small image set depicting the target concept, Textual Inversion~\cite{gal2022image} learned a new pseudo-word (\ie, S*) in the well-editable textual word embedding space of text encoder to represent the user-defined concept.
DreamBooth~\cite{ruiz2022dreambooth} finetuned the entire diffusion model to accurately align the target concept with a unique identifier.
Custom Diffusion~\cite{kumari2022multi} balanced the fidelity and memory by selectively finetuning K, V mapping parameters in cross attention layers.

Albeit flexible generation has been achieved by~\cite{gal2022image, ruiz2022dreambooth, kumari2022multi}, the computational efficiency remains a challenge to obtain the textual embedding of a visual concept.
Existing methods usually adopt the per-concept optimization formulation, which requires several or tens of minutes to learn a single concept.
As shown in Fig.~\ref{fig:comparison}, the Custom Diffusion~\cite{kumari2022multi}, which is among the fastest existing algorithms, still takes around 6 minutes to learn one concept, which is infeasible for online applications.
In contrast, in GAN inversion, many efficient learning-based methods~\cite{richardson2021encoding} have been proposed to accelerate the optimization process. 
An encoder can be trained to infer the latent codes, which only needs one step forward inference.

\begin{figure}[t]
\centering
\includegraphics[width=1\linewidth]{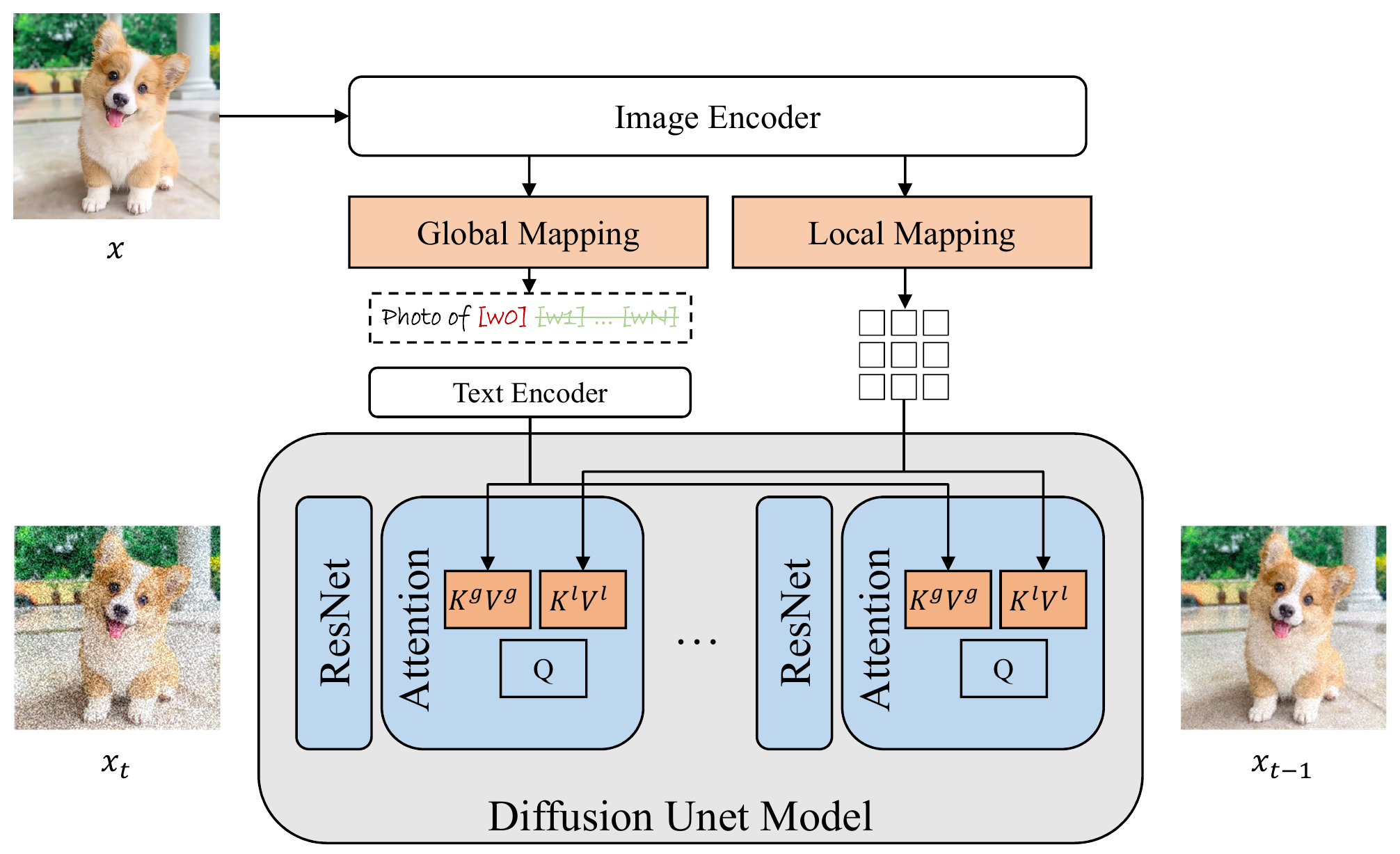}
\caption{\textbf{Inference pipeline of our proposed ELITE}.
Given a user-provided image $x$, our ELITE extracts hierarchical features with CLIP image encoder.
Then, it uses global and local mapping networks to encode the visual concept into textual word embeddings (\ie, primary word {\color{red} $w_{0}$} and auxiliary words {\color{green}$w_{1 \cdots N}$}) and textual feature embeddings, respectively.
The embeddings are injected with cross attention to guide customized generation.
Note that, in generation, only \textcolor{red}{$w_0$} is used for better editability.
}
\label{fig:inference}
\vspace{-1em}
\end{figure}

Driven by the above analysis, we propose a learning-based encoder for \textbf{E}ncoding visua\textbf{L} concepts \textbf{I}nto \textbf{T}extual \textbf{E}mbeddings, termed as \textbf{ELITE}. 
As shown in Fig.~\ref{fig:inference}, our ELITE adopts a pre-trained CLIP image encoder~\cite{radford2021learning} for feature extraction, followed by a global mapping network and a local mapping network to encode visual concepts into textual embeddings.
Firstly, we train a global mapping network to map the CLIP image features into the textual word embedding space of the CLIP text encoder, which has superior editing capacity~\cite{gal2022image}.
Since a given image contains both the subject and irrelevant disturbances, encoding them as a single word embedding severely degrades the editability of subject concept.
Thus, we propose to separately learn them with a well-editable primary word and several auxiliary words.
Using the hierarchical features from CLIP intermediate layers, the word learned from the deepest features naturally links to the primary concept (\ie, the subject), while auxiliary words learned from other features describe the irrelevant disturbances (as shown in Fig.~\ref{fig:word_visualization}).
When deploying to customized generation, we only use the primary word to avoid editability degradation from auxiliary words.

Usually, a visual concept is worth more than one word, and describing it with a single word may result in the inconsistency of local details~\cite{gal2022image}.
For higher fidelity of the learned concept without sacrificing its editability, we further propose a local mapping network to inject finer details.
From Fig.~\ref{fig:inference}, our local mapping network encodes the CLIP features into the textual feature space (\ie, the output space of the text encoder).
Compared with the textual word embeddings learned by global mapping network, the textual feature embeddings focus on the local details of each patch in the given image.
Then, the obtained textual feature embeddings are injected through additional cross attention layers, and the output feature is fused with the global part to improve the local details.
Experiments show that our ELITE can encode the target concept efficiently and faithfully, while keeping control and editing abilities.
The contributions of this work are summarized as follows:
\vspace{-0.2em}
\begin{itemize}
    \setlength{\itemsep}{5pt}
    \setlength{\parsep}{0pt}
    \setlength{\parskip}{0pt}
    \item We propose a learning-based encoder, namely ELITE, for fast and accurate customized text-to-image generation. It adopts a global and a local mapping networks to encode visual concepts into textual embeddings.
    \item Multi-layer features are adopted in global mapping to learn a well-editable primary word embedding, while the local mapping improves the consistency of details without sacrificing editability.
    \item Experimental results show that our ELITE can faithfully recover the target concept with higher visual fidelity, and enable more robust editing.
\end{itemize}

\section{Related Work}

\subsection{Text-to-Image Generation}

Deep generative models have achieved tremendous success on text-conditioned image generation~\cite{balaji2022ediffi,saharia2022photorealistic,ramesh2022hierarchical,ding2021cogview,ding2022cogview2,gafni2022make,li2019controllable,nichol2021glide,ramesh2021zero,xia2021tedigan,chang2023muse,yu2022scaling,sauer2023stylegan,rombach2022high} and have recently attracted intensive attention.
They can be categorized into three groups: GAN-based, VAE-based, and diffusion-based models.
Albeit GAN-based~\cite{li2019controllable,xia2021tedigan,sauer2023stylegan} and VAE-based models~\cite{ding2021cogview,gafni2022make,ramesh2021zero,chang2023muse,yu2022scaling} can synthesize images with promising quality and diversity, they still cannot match user descriptions very well.
Recently, diffusion models have shown unprecedentedly high-quality and controllable imaginary generation and been broadly applied to text-to-image generation~\cite{balaji2022ediffi,saharia2022photorealistic,ramesh2022hierarchical,ding2022cogview2,nichol2021glide,rombach2022high}.
By training with massive corpora, these large text-to-image diffusion models, such as DALLE-2~\cite{ramesh2022hierarchical}, Imagen~\cite{saharia2022photorealistic}, and Stable Diffusion~\cite{rombach2022high} have demonstrated excellent semantic understanding, and can generate diverse and photo-realistic images according to a given text prompt.
However, despite the superior performance on general synthesis, they still struggle to express the specific or user-defined concepts, \eg, ``corgi'' in Fig.~\ref{fig:comparison}.
Our method focuses on making pre-trained diffusion models to learn these new concepts efficiently.

\begin{figure*}[t]
\centering
\includegraphics[width=1\linewidth]{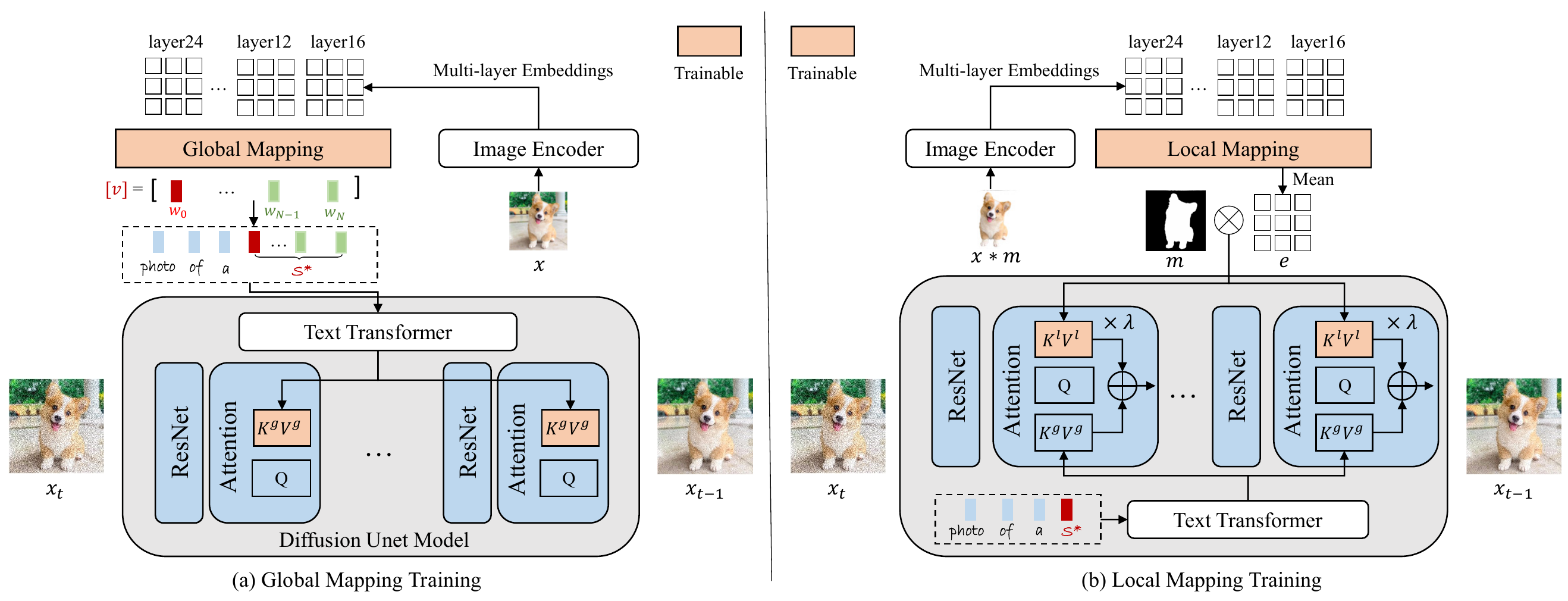}
\vspace{-1.5em}
\caption{\textbf{Training pipeline of our proposed ELITE}. Our method consists of two stages: \textbf{(a)} a global mapping network is first trained to encode a concept image into multiple textual word embeddings, with one primary word ({\color{red} $w_0$}) for well-editable concept and other auxiliary words ({\color{green} $w_{1 \cdots N}$}) to exclude irrelevant disturbances. \textbf{(b)} A local mapping network is further trained, which projects the foreground object into textual feature space to provide local details. In this stage, only the well-editable primary word ({\color{red} $w_0$}) is used.}
\label{fig:framework}
\vspace{-1.5em}
\end{figure*}

\subsection{GAN Inversion}

GAN inversion refers to projecting real images into latent codes so that images can be faithfully reconstructed and edited with pre-trained GAN models~\cite{xia2022gan, liu2022survey}.
Generally speaking, there are two types of GAN inversion algorithms in the literature: i) \textit{optimization-based:} directly optimize latent code to minimize the reconstruction error~\cite{lipton2017precise,bojanowski2017optimizing,creswell2018inverting}, and ii) \textit{encoder-based:} train an encoder to invert an image into latent space~\cite{parmar2022spatially,wang2022high,abdal2019image2stylegan,abdal2020image2stylegan++,richardson2021encoding}.
The optimization-based methods~\cite{lipton2017precise,bojanowski2017optimizing,creswell2018inverting} usually require hundreds of iterations to obtain promising results, while encoder-based methods greatly accelerate this process via one feed-forward pass only.
To improve image fidelity without compromising editability, HFGI~\cite{wang2022high} embeds the omitted information into high-rate features.
Similarly, our ELITE adopts a local mapping network that encodes the concept images into textual feature space to improve details consistency.

\subsection{Diffusion-based Inversion}

The inversion of text-to-image diffusion models can be performed in two types of latent spaces: the Textual Word Embedding (TWE) space~\cite{gal2022image,ruiz2022dreambooth,kumari2022multi,dong2022dreamartist} of text encoder or the Image-based Noise Map (INM) space~\cite{choi2021ilvr,song2020denoising,mokady2022null}.
INM-based inversion methods, such as DDIM~\cite{song2020denoising} or Null Text~\cite{mokady2022null} find the initial noise to reconstruct image faithfully, while it suffers from a degraded editing ability.
In contrast, the TWE space has shown superior editing capacity, which is well suitable for textual inversion and customized generation~\cite{gal2022image, kumari2022multi, dong2022dreamartist}.
For example, Textual Inversion~\cite{gal2022image} and DreamArtist~\cite{dong2022dreamartist} optimize the embedding of new ``words'' using a few user-provided images to recover the target concept.
DreamBooth~\cite{ruiz2022dreambooth} finetunes the entire text-to-image model to learn high-fidelity new concept with a unique identifier.
To improve the computation efficiency, Custom Diffusion~\cite{kumari2022multi} only updates the key and value mapping parameters in cross attention layers with better performance.
Albeit flexible generation has been achieved, existing methods still require several or tens of minutes to learn a single concept.

In this work, we choose the TWE space as target for inversion, while proposing a learning-based encoder ELITE for fast and accurate customized text-to-image generation.
With the proposed global and local mapping networks, our method can learn the new concept quickly and faithfully with one single image.

\vspace{-0.5em}
\section{Proposed Method}
\vspace{-0.5em}

Given a pretrained text-to-image model $\epsilon_\theta$ and an image $x$ indicating the target concept (usually an object), customized text-to-image generation aims to learn a pseudo-word (S*) in word embedding space to describe the concept faithfully, while keeping the editability.
To achieve fast and accurate customized text-to-image generation, we propose an encoder ELITE to encode the visual concept into textual embeddings.
As illustrated in Fig.~\ref{fig:inference}, our ELITE first adopts a global mapping network to encode visual concepts into the textual word embedding space.
The obtained word embeddings can be composed with texts flexibly for customized generation.
To address the information loss in word embedding, we further propose a local mapping network to encode the visual concept into textual feature space to improve the consistency of the local details.
In the following, we begin by presenting an overview of the text-to-image model utilized in our approach (Sec.~\ref{sec:pre}).
Then, we will introduce the details of the proposed global mapping network (Sec.~\ref{sec:global}) and local mapping network(Sec.~\ref{sec:local}).

\begin{figure*}[t]
\centering
\includegraphics[width=1\linewidth]{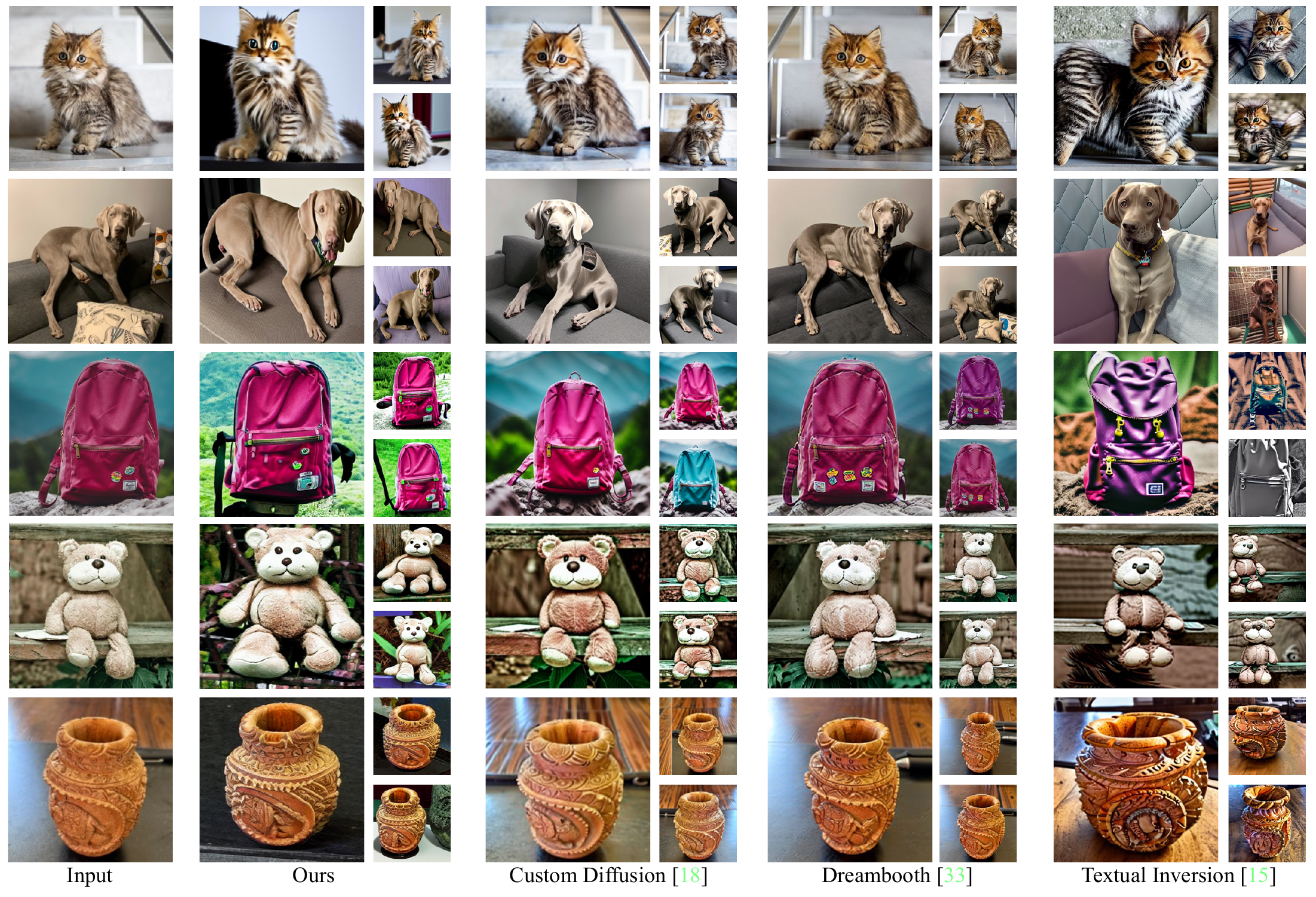}
\vspace{-1.2em}
\caption{\textbf{Visual comparisons of concept generation.} For generating images, \texttt{A photo of a S*} is used for Textual Inversion~\cite{gal2022image} and our method, while \texttt{A photo of a S* [category]} is used for Dreambooth~\cite{ruiz2022dreambooth} and Custom Diffusion~\cite{kumari2022multi}. Our ELITE shows comparable performance to it competitors.}
\label{fig:inversion}
\vspace{-1.5em}
\end{figure*}

\subsection{Preliminary}
\label{sec:pre}

In this work, we employ the Stable Diffusion~\cite{rombach2022high} as our
text-to-image model, which is trained on large-scale data and consists of two components.
First, the autoencoder ($\mathcal{E}(\cdot)$, $\mathcal{D}(\cdot)$) is trained to map an image $x$ to a lower dimensional latent space by the encoder $z = \mathcal{E}(x)$.
While the decoder $\mathcal{D}(\cdot)$ learns to map the latent code back to the image so that $ D(\mathcal{E}(x)) \approx x $.
Then, the conditional diffusion model $\epsilon_{\theta}(\cdot)$ is trained 
on the latent space to generate latent codes based on text condition $y$.
We simply adopt the mean-squared loss to train the diffusion model:
\begin{equation}
\small
L_{LDM} := \mathbb{E}_{z\sim\mathcal{E}(x), y, \epsilon \sim \mathcal{N}(0, 1), t }\Big[ \Vert \epsilon - \epsilon_\theta(z_{t},t, \tau_\theta(y)) \Vert_{2}^{2}\Big] \, ,
\label{eq:LDM_loss}
\end{equation}
where $\epsilon$ denotes the unscaled noise, $t$ is the time step, $z_t$ is the latent noise at time $t$, and $\tau_\theta(\cdot)$ represents the pretrained CLIP text encoder~\cite{radford2021learning}.
During inference, a random Gaussian noise $z_T$ is iteratively denoised to $z_0$, and the final image is obtained through the decoder $x' = \mathcal{D}(z_0)$.

To incorporate text information in the process of image generation, cross attention is adopted in Stable Diffusion.
Specifically, the latent image feature $f$ and text feature $\tau_\theta(y)$ are first transformed by the projection layers to obtain the query $Q = W_Q \cdot f$, key $K = W_K \cdot \tau_\theta(y)$ and value $V = W_V \cdot \tau_\theta(y)$.
$W_Q$, $W_K$, and $W_V$ are weight parameters of query, key, and value projection layers, respectively.
Attention is conducted by a weighted sum over value features:
\begin{equation}
\small
\text{Attention}(Q, K, V) = \text{Softmax}\left( \frac{QK^T}{\sqrt{d'}} \right) V,
\label{eq:attention}
\end{equation}
where $d'$ is the output dimension of key and query features.
The latent image feature is then updated with the output of the attention block.

\subsection{Global Mapping}
\label{sec:global}

Following~\cite{gal2022image, kumari2022multi}, we choose the textual word embedding space of CLIP text encoder as the target for inversion.
To improve the computation efficiency, we propose a global mapping network that encodes the given concept image into word embeddings directly.
As illustrated in Fig.~\ref{fig:framework}(a), to facilitate the embedding learning, the pretrained CLIP image encoder $\psi_\theta(\cdot)$ is adopted as feature extractor, and our global mapping network $M^g(\cdot)$ projects the CLIP features as word embeddings $v$:
\begin{equation}
    v = M^g \circ \psi_\theta(x),
\end{equation}
where $v \in \mathbb{R}^{N\times d}$, $N$ is the number of words and $d$ is the dimension of word embedding.
Global average pooling is employed on features to obtain the word embedding.

\begin{figure}[t]
\centering
\includegraphics[width=1\linewidth]{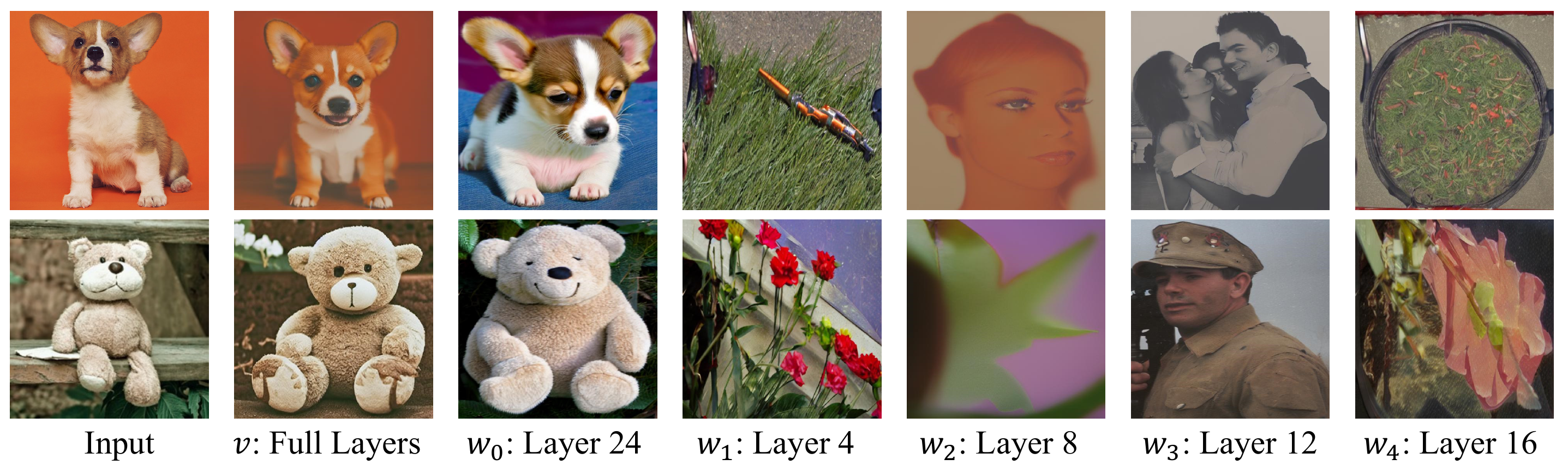}
\caption{\textbf{Visualization of learned word embeddings}.
The word associated with the deepest feature (\ie, $w_0$ from layer 24) describes the primary concept (\ie, corgi, teddybear), while other words describe irrelevant disturbances.}
\label{fig:word_visualization}
\vspace{-1.2em}
\end{figure}

Since the image $x$ contains both the desired subject and irrelevant disturbances, encoding them into one word (\ie, $N=1$) results in an entangled word embedding $v$ with poor editability.
To obtain a more informative and editable word embedding, we adopt a multi-layer approach to learn $N$ ($N$ \textgreater $1$) words from the image $x$ separately.
Specifically, we select $N$ layers from the CLIP image encoder, and each layer $\psi_\theta^{Li}(\cdot)$ learns one word $w_i$ independently.
All words $[w_0, \cdots, w_N]$ are concatenated together to form the textual word embedding $v$.
A pseudo-word S* is further introduced to represent the learned new concept, and $v$ is associated with its word embeddings.
To train our global mapping network, we adopt Eqn.~(\ref{eq:LDM_loss}), and regularize the obtained word embeddings as follows:
\begin{equation}
    L_{global} = L_{LDM} + \lambda_{global} \Vert v \Vert_1,
\end{equation}
where $\lambda_{global}$ is a trade-off hyperparameter.
Analogous to~\cite{gal2022image}, we randomly sample a text from the CLIP ImageNet templates~\cite{radford2021learning} as text input during training, such as \texttt{a photo of a S*}. 
The full template list is provided in the \emph{Suppl}.
Besides, following~\cite{kumari2022multi}, the key and value projection layers in the cross attention layer are finetuned with $M^g(\cdot)$, and the obtained new projections are denoted as $K^g = W^g_K \cdot \tau_\theta (y)$ and $V^g = W^g_V \cdot \tau_\theta (y)$.

Benefiting from the hierarchical semantics learned by different layers in the CLIP image encoder, the feature from the deepest layer (\ie, layer 24) possesses the highest comprehension of the image.
The word embedding associated with the deepest feature is naturally learned to describe the primary concept (\ie, the subject), while keeping superior editability.
In contrast, word embeddings from shallower features are learned to describe the irrelevant disturbances (see Sec.~\ref{sec:abla} for more details).
Note that, we use only the word embedding of the deepest feature during local training (Sec.~\ref{sec:local}) and image generation stages for better editability.

\begin{figure*}[t]
\centering
\includegraphics[width=1\linewidth]{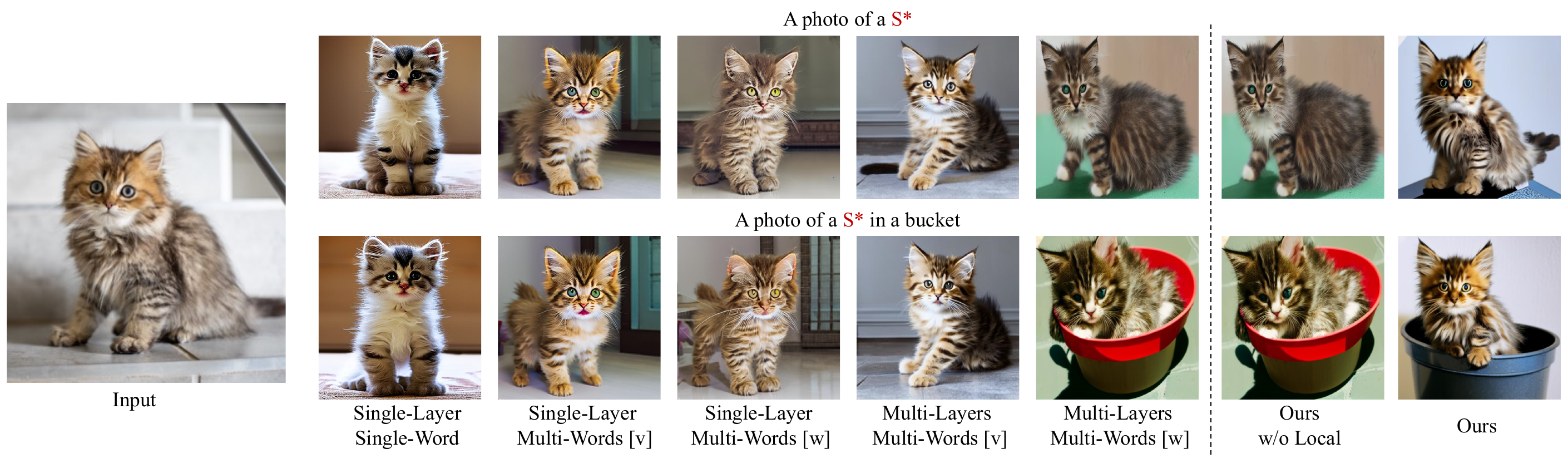}
\vspace{-1.5em}
\caption{\textbf{Visual comparisons of different variants.} \textbf{Left}: [v] denotes the generation results with full word embeddings $v$, while [w] denotes the generation results with the primary word embedding $w$. Learning single or multiple word embeddings from the single layer (\ie the deepest layer) fail to achieve reliable editing capacity. In contrast, learning multiple words from multiple layers (\ie, our method) successfully learns an editable primary word embedding. \textbf{Right}: Our proposed local mapping significantly improves the consistency of details between the input image and the generated image, while maintains editability. }
\label{fig:ablation}
\vspace{-0.5em}
\end{figure*}

\begin{figure*}[t]
\centering
\includegraphics[width=1\linewidth]{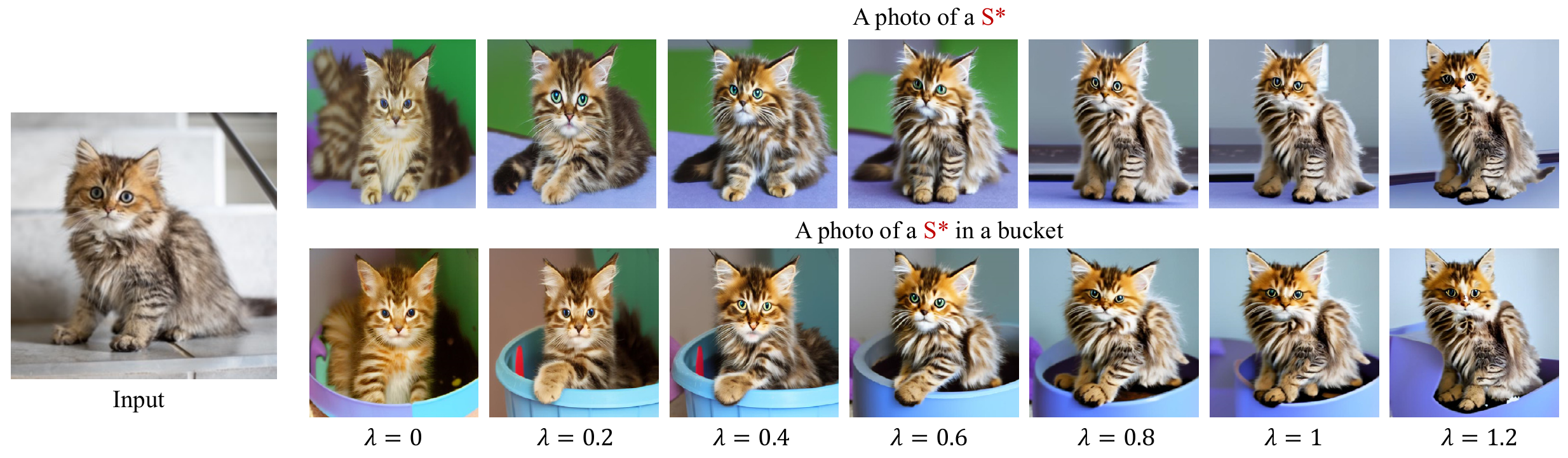}
\vspace{-1.5em}
\caption{\textbf{Visual comparisons by using different values of $\lambda$}. As $\lambda$ increases, the consistency of details between the generated image and the input image improves, yet slightly reducing the editability.}
\label{fig:lambda_ablation}
\vspace{-1.5em}
\end{figure*}

\subsection{Local Mapping}
\vspace{-0.4em}
\label{sec:local}

Usually, a single word embedding is not sufficient to faithfully describe the details of the given concept, while multiple word embeddings may suffer from degraded editing capacity.
To improve the consistency between the given concept and synthesized image without sacrificing editability, we further propose a local mapping network.
As shown in Fig.~\ref{fig:framework}(b), the local mapping network 
$M^l(\cdot)$ encodes the multi-layer CLIP features into the textual feature space (\ie, the output space of text encoder):
\begin{equation}
    e = M^l \circ \psi_\theta(x * m),
\end{equation}
where $m$ is the object mask to ease the redundant details of background.
$e \in \mathbb{R}^{p \times p \times d}$ keeps the spatial structure and $p$ is the feature size.
Each pixel of $e$ focuses on the local details of each patch in the given image.
To inject the local information of $e$ into generation, we introduce two additional projection layers $W^l_K$ and $W^l_V$ into cross attention module.
The local attention is performed by $\text{Attention} (Q, K^l, V^l)$, where $K^l = W^l_K \cdot (e * m)$ and $V^l = W^l_V \cdot (e * m)$.
Then, the output feature of local attention is fused with global part to inject finer details:
\begin{equation}
\small
\textit{Out} = \text{Attention} (Q, K^g, V^g) + \lambda \text{Attention} (Q, K^l, V^l), 
\label{eq:fuse}
\end{equation}
where $\lambda$ is a hyperparameter and set as $1$ during training.
To emphasize more on the object region, the obtained attention map $Q{K^l}^{T}$ is reweighted by $Q{K^g}^{T}_i$, where $i$ is the index of $w_0$ in the text prompt (more details will be provided in the \textit{Suppl}).
To train the local mapping network, we also adopt Eqn.~(\ref{eq:LDM_loss}) while regularizing the local values $V_l$:
\begin{equation}
    L_{local} = L_{LDM} + \lambda_{local} \Vert V^l \Vert_1,
\end{equation}
where $\lambda_{local}$ is a trade-off hyperparameter.

\begin{table}[t]
\begin{center}
\caption{\textbf{Ablation study}. [v] denotes the generated testing results with full word embeddings $v$, while [w] denotes the generated testing results with the primary word embedding $w$.}
\label{tab:ablation}
\resizebox{1\linewidth}{!}{
    \setlength{\tabcolsep}{4mm}
    \begin{tabular}{lccc}
        \toprule
        Method & CLIP-T ($\uparrow$) & CLIP-I ($\uparrow$) & DINO-I ($\uparrow$) \\
        \midrule
        Single-Layer Single-Word & 0.209 & 0.714 & 0.546  \\
        Single-Layer Multi-Words [v] & 0.198 & 0.683 & 0.431  \\
        Single-Layer Multi-Words [w]  & 0.212 & 0.692 & 0.443 \\
        Multi-Layers Multi-Words [v]  & 0.204 & \textbf{0.771} & \textbf{0.658}  \\
        Multi-Layer Multi-Words [w]  & \textbf{0.257} & 0.699 & 0.486  \\
        \midrule
        Ours w/o Local & \textbf{0.257} & 0.699 & 0.486  \\
        Ours & 0.255 & 0.762 & 0.652  \\
        \bottomrule
    \end{tabular}
}
\end{center}
\vspace{-2.2em}
\end{table}

\section{Experiments}

\subsection{Experimental Settings}

\noindent \textbf{Datasets.} 
To train our local and global mapping networks, we use the \texttt{testset} of OpenImages~\cite{kuznetsova2020open} as our training dataset. 
It contains 125k images with 600 object classes. 
During training, we crop and resize the object image to 512$\times$512 according to the bounding box annotations.
While for local mapping training, mask annotations are also used to extract foreground objects.
For customized generation, we adopt concept images from existing works~\cite{gal2022image, ruiz2022dreambooth, kumari2022multi} with 20 subjects, including dog, cat, and toy, \etc. 
The subject masks can be obtained by a pretrained segmentation model~\cite{zhang2022mining, kirillov2023segment, chen2022pp}.
For quantitative evaluation, we employ the editing prompts from~\cite{ruiz2022dreambooth}, which contains 25 editing prompts for each subject. 
We randomly generate five images for each subject-prompt pair, obtaining 2,500 images in total.
More details can be found in the \textit{Suppl}.

\noindent \textbf{Evaluation metrics.} 
Following Dreambooth~\cite{ruiz2022dreambooth}, we evaluate our method with three metrics: \textit{CLIP-I}, \textit{CLIP-T}, and \textit{DINO-I}.
For \textit{CLIP-I}, we calculate the CLIP visual similarity between the generated and target concept images.
For \textit{CLIP-I}, we calculate the CLIP text-image similarity between the generated images and the given text prompts.
The pseudo-word (S*) in the text prompt is replaced with the proper object category for extracting CLIP text feature.
For \textit{DINO-I}, we calculate cosine similarity between the ViTS/16 DINO~\cite{caron2021emerging} embeddings of generated and concept images.
Moreover, we adopt the \textit{optimization time} as a metric to evaluate the efficiency of each method.

\noindent \textbf{Implementation Details.} 
We use the V1-4 version of Stable Diffusion in our experiments, and the mapping network is implemented with three-layer MLP (for both global mapping network and local mapping network). 
To extract multi-layer CLIP features,  features from five layers are selected, and the layer indexes are $\{24, 4, 8, 12, 16\}$ in order.
To train the global mapping network, we use the batch size of 16 and $\lambda_{global} = 0.01$. 
The learning rate is set to 1e-6.
To train the local mapping network, we adopt the batch size of 8 and $\lambda_{local} = 0.0001$.
The learning rate is set to 1e-5.
All experiments are conducted on 4$\times$V100 GPUs.
During image generation, we use 100 steps of the LMS sampler, and the scale of classifier-free guidance is 5. 
Unless mentioned otherwise, we use $\lambda = 0.8$ for concept generation (\eg, \texttt{a photo of a S*}), and $\lambda = 0.6$ for concept editing (\eg, \texttt{a S* wearing sunglasses}).

\subsection{Ablation Study}
\label{sec:abla}

We first conduct the ablation studies to evaluate the effects of various components in our method, including the multi-layer features in the global mapping network, local mapping network, and the value of $\lambda$.

\noindent \textbf{Effect of Multi-layer Features.}
Fig.~\ref{fig:word_visualization} gives the visualization of words learned by multi-layer features in the global mapping network. 
For each word visualization, we use the text \texttt{A photo of a [$w_i$]}.
One can see that, the word embedding of the deepest feature (\ie, $w_0$) describes the primary concept (\ie, corgi, teddybear), while other words describe some irrelevant details.
Meanwhile, the obtained $w_0$ maintains superior editability.
To further demonstrate this, we conducted experiments with several variants:
i) Single-layer Single-word: learning a single word embedding from the deepest feature.
ii) Single-layer Multi-words: learning multiple word embeddings from the deepest feature separately.
iii) Multi-layers Multi-words: our setting, learning multiple word embeddings from the multiple layer features separately.
Fig.~\ref{fig:ablation} illustrates the results of concept generation and editing for each variant.
For multiple word settings, we show the results of the full embeddings (\ie, [$v$] ) and the primary word embedding (denoted as [$w$]).
As shown in the figure, encoding concept image into one single word embedding leads to entangled embedding with poor editability.
When learning multiple words from the deepest feature, the obtained full word embeddings $v$ and the primary word embedding $w$ are not editable either.
In contrast, the primary word $w$ learned by our multi-features describes the object concept while maintaining superior editing capacity.
That's why we only keep it during image generation.
Since a single $w$ is not sufficient to describe the details of the given concept faithfully, a local mapping network is further proposed to address this.

\begin{table}[!t]
\centering
\caption{\textbf{Quantitative comparisons with existing methods.}}
\setlength{\tabcolsep}{10pt}
\resizebox{\linewidth}{!}{
\begin{tabular}{l|ccc|c}
\toprule
Method & CLIP-T ($\uparrow$) & CLIP-I ($\uparrow$) &  DINO-I ($\uparrow$) & Time ($\downarrow$) \\
\midrule
Textual Inversion~\cite{gal2022image}  & 0.183 & 0.663 & 0.462 & 50 min  \\
DreamBooth~\cite{ruiz2022dreambooth} & 0.251  & 0.785 & 0.674 & 15 min \\
Custom Diffusion~\cite{kumari2022multi} & 0.245 & \textbf{0.801}  & \textbf{0.695} & 6 min \\
\midrule
Ours & \textbf{0.255} & 0.762 & 0.652 & \textbf{0.05s} \\
\bottomrule
\end{tabular}
}
\label{tab:results}
\end{table}

\begin{table}[t]
\begin{center}
\caption{\textbf{User study.} The numbers indicate the percentage (\%) of volunteers who favor the results of our method over those of the competing methods based on the given question.}
\vspace{-0.6em}
\label{tab:user_study}
\resizebox{1\linewidth}{!}{
    \renewcommand{\arraystretch}{1.2}
    \begin{tabular}{lccc}
        \toprule
        Metric & \makecell[c]{Ours vs. \\ Textual Inversion} & \makecell[c]{Ours vs. \\ Dreambooth}  & \makecell[c]{Ours vs. \\ Custom Diffusion}  \\
        \midrule
        Text-alignment & 75.09 & 57.09 & 62.77 \\
        Image-alignment & 86.59 & 45.50 & 43.77 \\
        Editing-alignment & 90.18 & 52.09 & 48.18 \\
        \bottomrule
    \end{tabular}
}
\end{center}
\vspace{-2em}
\end{table}

\noindent \textbf{Effect of Local Mapping.}
We further conduct the ablation to evaluate the effect of the proposed local mapping network.
As shown in Fig.~\ref{fig:ablation}, with the local mapping network, our ELITE generates images with higher consistency with the concept image.
Meanwhile, from Table~\ref{tab:ablation} we see that the introduction of a local mapping network does not compromise the editable capability, demonstrating its superiority over learning multiple words.
Though its image alignment may not be the best, it has a good trade-off between image alignment and text alignment.

\noindent \textbf{Effect of $\lambda$.}
In Eqn.~\ref{eq:fuse}, $\lambda$ is introduced to control the fusion of information from the global mapping network and the local mapping network.
To evaluate its effect, we vary its value from 0 to 1.2, and the generated results are shown in Fig.~\ref{fig:lambda_ablation}.
We see that with the increase of $\lambda$, the consistency between the synthesized image and concept image is improved.
However, when the value of $\lambda$ is too large, it may lead to degenerated editing results. 
Therefore, for a good trade-off between inversion and editability, we set $\lambda=0.6$ for editing prompts and $\lambda=0.8$ for generating prompts.
We find these parameters work well for most cases.

More ablations are provided in the \emph{Suppl}.

\begin{figure*}[t]
\centering
\includegraphics[width=1\linewidth]{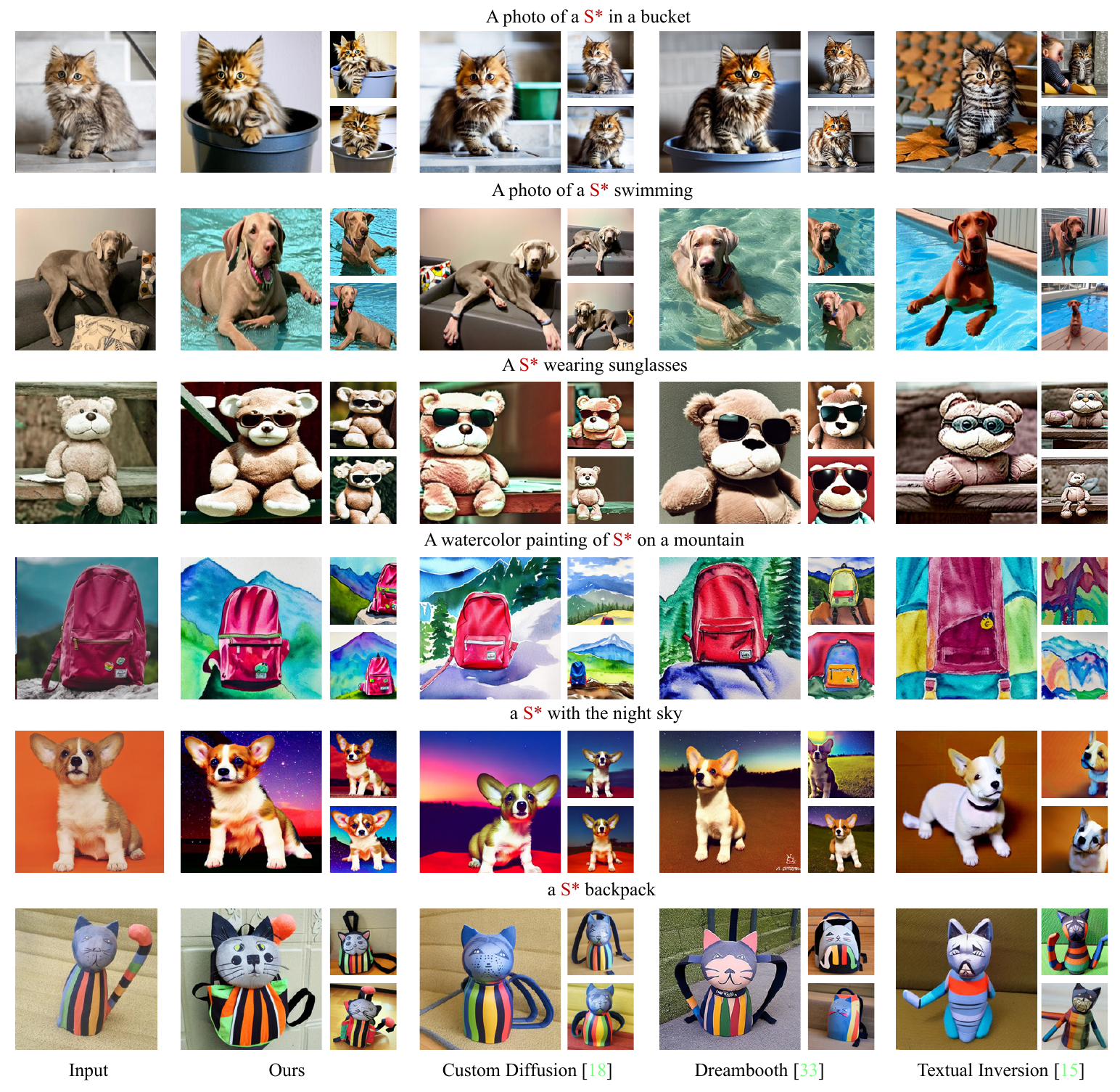}
\vspace{-1.5em}
\caption{\textbf{Visual comparisons of concept editing.} Our ELITE method demonstrates superior editability compared to Textual Inversion~\cite{gal2022image}, Dreambooth~\cite{ruiz2022dreambooth}, and Custom Diffusion~\cite{kumari2022multi}. }
\label{fig:editing}
\vspace{-0.5em}
\end{figure*}

\begin{figure}[t]
\centering
\vspace{-0.6em}
\includegraphics[width=1\linewidth]{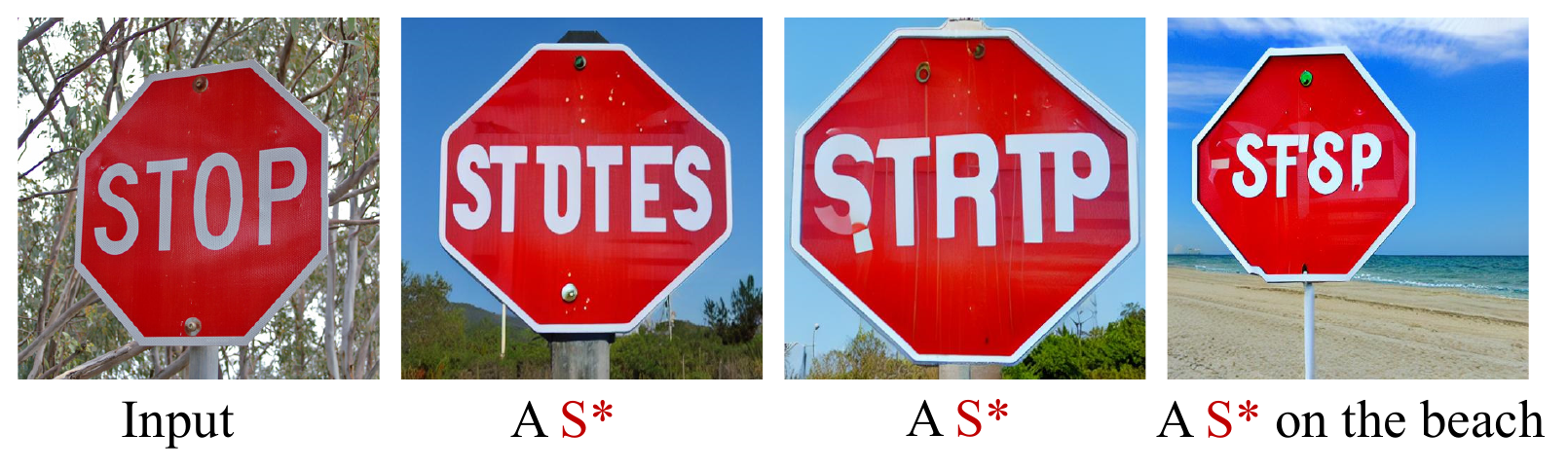}
\caption{\textbf{Failure cases.} Similar to stable diffusion, our method fails to deal with images involving text characters.}
\label{fig:failure_case}
\vspace{-1em}
\end{figure}

\subsection{Qualitative Results}

\begin{figure*}[t]
\centering
\includegraphics[width=1\linewidth]{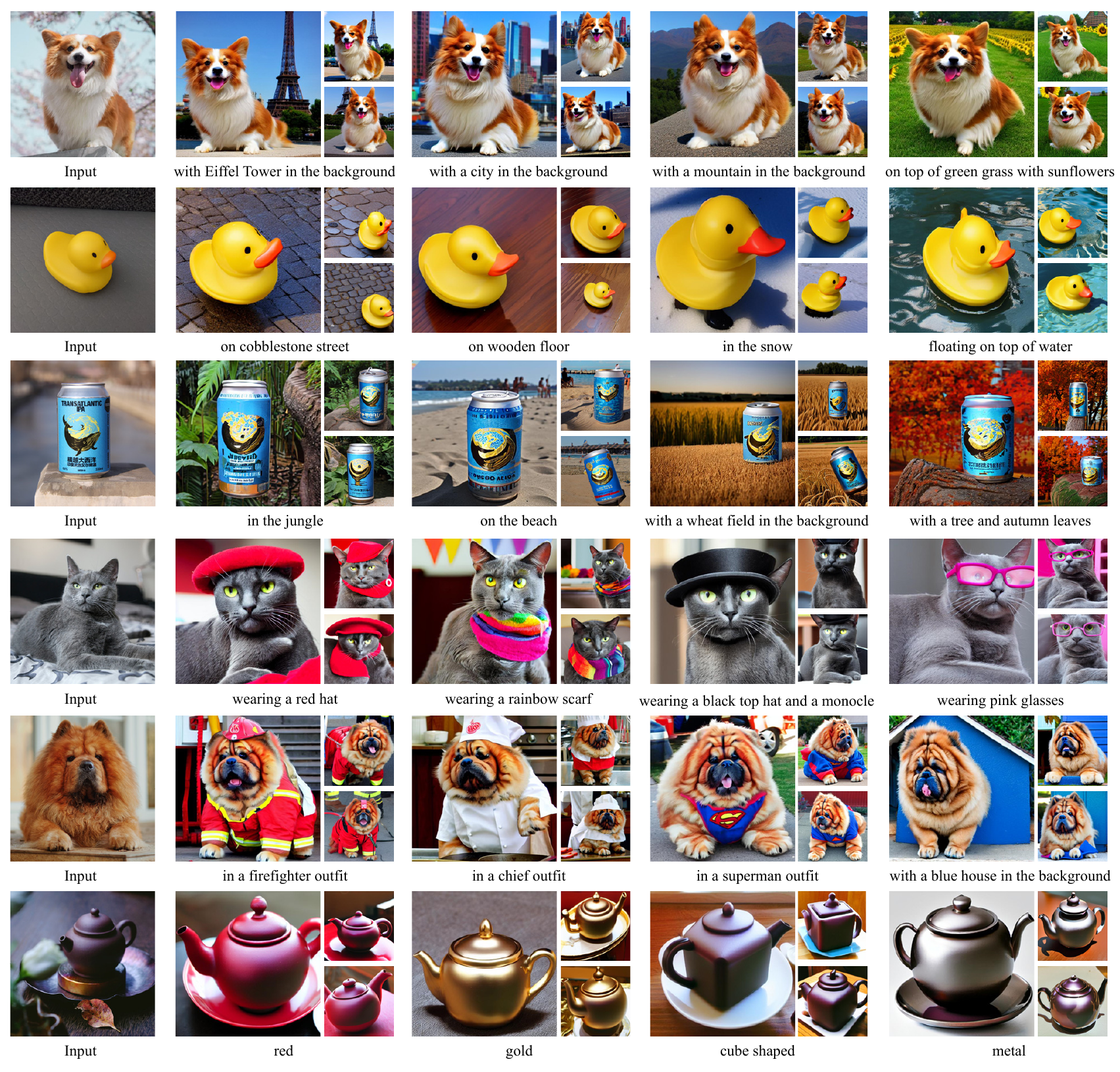}
\vspace{-0.5em}
\caption{\textbf{More visualization results.} Our ELITE can generate subjects with different contexts, accessories and properties consistently.}
\label{fig:more_results}
\vspace{-0.5em}
\end{figure*}

To demonstrate the effectiveness of our ELITE, we compare it with existing optimization-based methods, including Textual Inversion~\cite{gal2022image}, DreamBooth~\cite{ruiz2022dreambooth}, and Custom Diffusion~\cite{kumari2022multi}.
For a fair comparison, we trained all models using their official codes\footnote{Since the official code of Dreambooth is not publicly available, we use the code implemented by \url{https://github.com/XavierXiao/Dreambooth-Stable-Diffusion}. For Textual Inversion, we use its stable diffusion version.} and default hyperparameters on a single image.
Fig.~\ref{fig:inversion} illustrates the images generated with the text prompt \texttt{A photo of a S*}.
With only one concept image, Textual Inversion cannot learn a word embedding to accurately describe the target concept.
Although Dreambooth and Custom Diffusion learn the concept with detail consistency, their diversity may be limited.
In comparison, our ELITE is capable of faithfully capturing the details of the target concept and generating diverse images. 
We also conduct evaluation with editing prompts and compare our method with existing methods.
As shown in Fig.~\ref{fig:editing}, Dreambooth and Custom Diffusion exhibit degraded editing ability, and in some cases, the editing prompts fail to produce the desired results (first row).
In contrast, our method demonstrates superior editing performance.
Fig.~\ref{fig:more_results} illustrates more qualitative results obtained by our method. 
One can see that our ELITE can generate various subjects with different contexts, accessories, and properties consistently, demonstrating its effectiveness.

\subsection{Quantitative Results}
In addition to the qualitative comparisons, we further conduct the quantitative evaluation to validate the performance of our ELITE. 
From Table~\ref{tab:results}, one can see that our method achieves better text-alignment compared to the state-of-the-art methods, demonstrating its superior editability. 
Moreover, our method achieves comparable detail consistency and image quality, indicating its capability to generate high-quality images.
Furthermore, our method provides a significant merit in terms of computational efficiency. 
Unlike optimization-based methods that require several or tens of minutes to obtain the concept embedding, our method can finish it in just 0.05s. 
This makes our method highly practical and efficient for real-world applications where speed is a critical factor.

\noindent \textbf{User Study.} 
We then perform the user study to compare with competing methods.
Given a subject, a text prompt and two synthesized images (ours \textit{v.s.} competitor), the users are asked to select the better one from three views:
i) Text alignment: ``Which image is more consistent with the text?''.
ii) Image alignment: ``Which image better represents the objects in target images?''.
iii) Editing alignment: ``Which image is more consistent with both the target subject and the text?''.
For each evaluated view, we employ 60 users, and each user is asked to answer 30 randomly selected questions, \ie, 1800 responses in total.
As shown in Table~\ref{tab:user_study}, our method receives comparable preference to others.

\subsection{Limitations}

As shown in Fig.~\ref{fig:failure_case}, our ELITE inherits the weakness from stable diffusion, 
\ie, failing to deal with images involving text characters.

\section{Conclusion}
In this paper, we proposed a novel learning-based encoder, namely ELITE, for fast and accurate customized text-to-image generation.
Compared with existing optimization-based methods, our ELITE directly encoded visual concepts into textual embeddings, significantly reducing the computational and memory burden of learning new concepts.
Moreover, our method demonstrates superior flexibility in editing the learned concepts into new scenes while preserving the image-specific details, making it a valuable tool for customized text-to-image generation.
In future work, we will explore to leverage multiple concept images for better inversion, and investigate effective methods for composing multiple concepts in ELITE.

\noindent \textbf{Acknowledgement.} This work was supported in part by National Key R\&D Program of China under Grant No. 2020AAA0104500, the National Natural Science Foundation of China (NSFC) under Grant No.s U19A2073 and 62006064, and the Hong Kong RGC RIF grant (R5001-18).

{\small
\bibliographystyle{ieee_fullname}
\bibliography{egbib}
}

\newpage

\renewcommand\thesection{\Alph{section}}
\renewcommand\thesubsection{\thesection.\arabic{subsection}}
\setcounter{section}{0}

\def\pholder{$S_*$}
\def\pholdercolor{{\color{blue}$S_*$}}
\def\mathpholder{S_*}
\def\mathpholdercolor{{\color{blue}S_*}}

\section{More Ablation Studies}
\label{sec:abla_more}

\subsection{Effect of the value of $\lambda$}
\begin{figure}[t]
\centering
\includegraphics[width=1\linewidth]{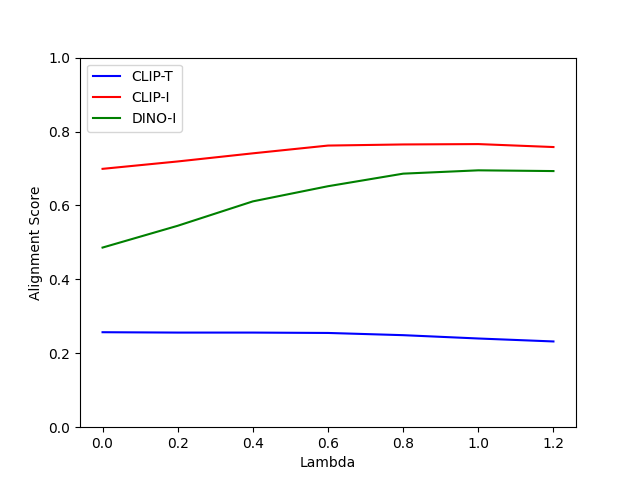}
\caption{Ablation on the value of $\lambda$. As $\lambda$ increases, the CLIP-I and DINO-I improve, yet slightly affects the text alignment (CLIP-T).}
\label{fig:lambda_ablation_quantitative}
\end{figure}

From Eqn.~({\color{red} 6}) in main paper, $\lambda$ is introduced to control the fusion of information from the global mapping network and the local mapping network.
To evaluate its effect, we vary its value from $0$ to $1.2$ during customized generation.
As shown in Fig.~{\color{red} 7} in main paper, with the increasing of $\lambda$, the consistency between the synthesized image and concept image is improved.
Meanwhile, from Fig.~\ref{fig:lambda_ablation_quantitative}, the image alignment (\ie, CLIP-I and DINO-I) improves as $\lambda$ increases.
However, when the value of $\lambda$ is too large, it may lead to degenerated editing results, resulting decreased text alignment (\ie, CLIP-T). 
Therefore, for a trade-off between inversion and editability, we set $\lambda=0.6$ for editing prompts and $\lambda=0.8$ for generating prompts.
We find these parameters work well for most cases.

\subsection{Effect of the layer indexes}

\begin{figure*}[t]
\centering
\includegraphics[width=1\linewidth]{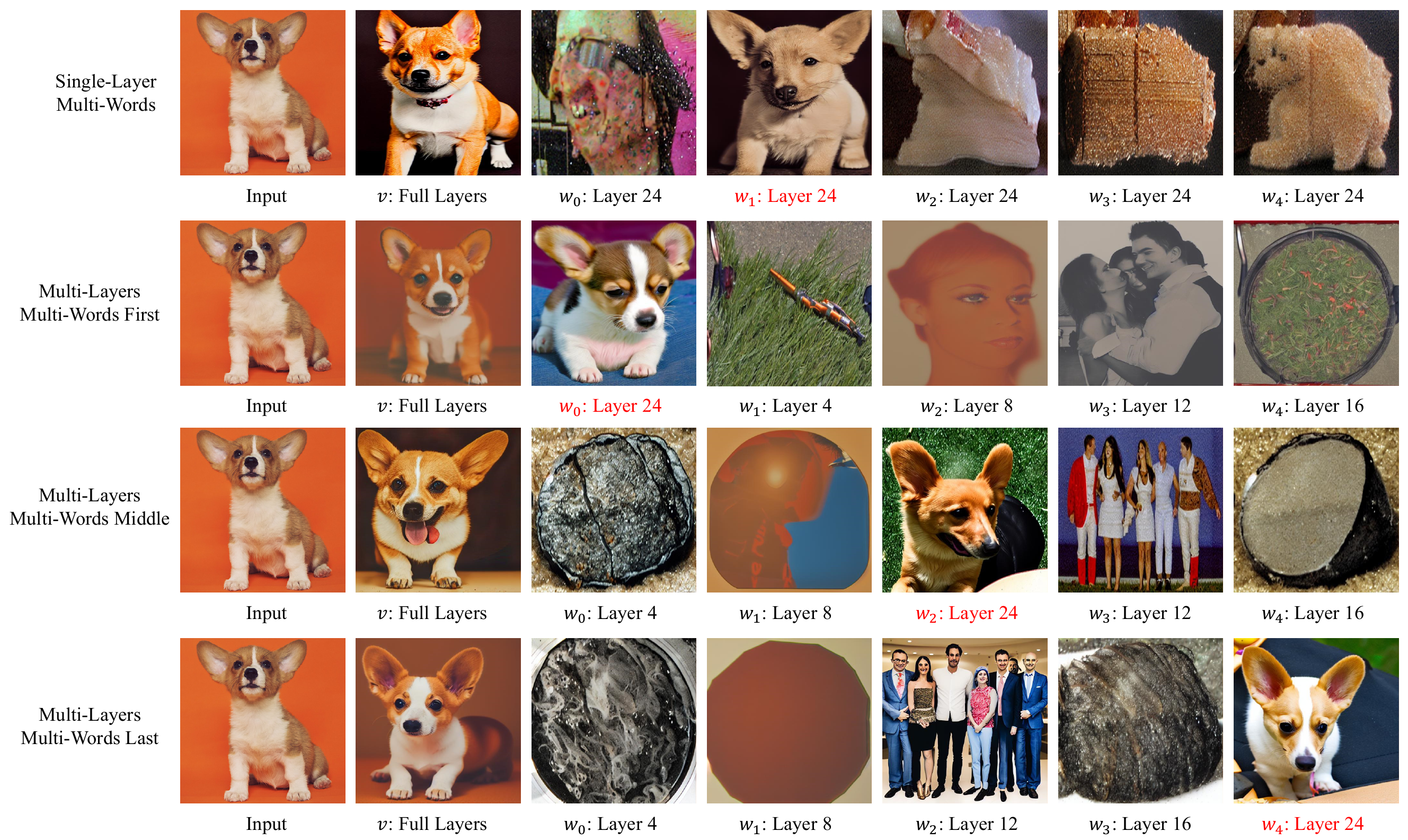}
\caption{\textbf{Visualization of learned word embeddings for different variants.} First, Middle, and Last denote the order of deepest feature in learned word embeddings. The learned primary word is highlighted by {\color{red}red} color. }
\label{fig:location_ablation}
\end{figure*}

\begin{figure*}[t]
\centering
\includegraphics[width=1\linewidth]{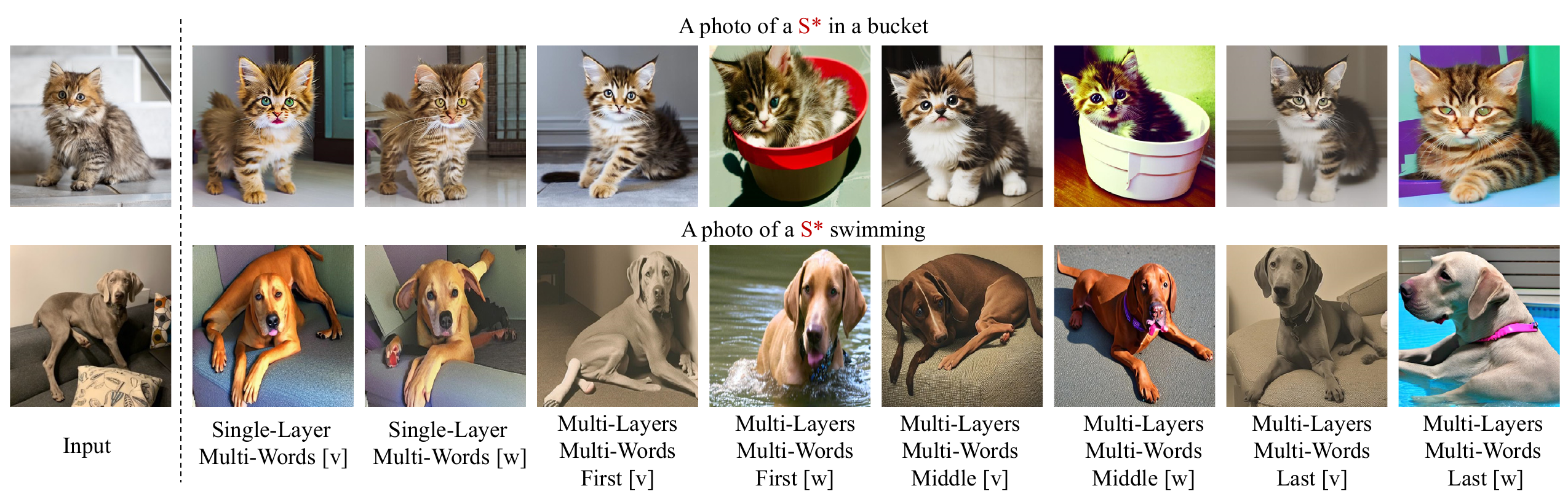}
\caption{\textbf{Visual comparisons of different variants.} [v] denotes the generation results with full word embeddings $v$, while [w] denotes the generation results with the primary word embedding $w$ (see Fig.~\ref{fig:location_ablation}). The primary word learned by multi-layer features is well-editable.}
\vspace{-1em}
\label{fig:location_ablation_editing}
\end{figure*}

In our experiments, we select the features of the five layers from CLIP image encoder to learn multiple word embeddings, whose indexes are \{$24, 4, 8, 12, 16$\} in order.
We have further conducted the ablation studies by putting the deepest layer (\ie, layer $24$) in different orders.
Specifically, we compare four variants, 
i) \textbf{Single-layer Multi-words}: learning multiple word embeddings from the deepest feature separately.
ii) \textbf{Multi-layers Multi-words First}: our setting, learning multiple word embeddings from the multiple layer features separately, and the layer indexes are \{$24, 4, 8, 12, 16$\} in order.
iii) \textbf{Multi-layers Multi-words Middle}: learning multiple word embeddings from the multiple layer features separately, and the layer indexes are \{$4, 8, 24, 12, 16$\} in order. 
iv) \textbf{Multi-layers Multi-words Last}: learning multiple word embeddings from the multiple layer features separately, and the layer indexes are \{$4, 8, 12, 16, 24$\} in order. 
Fig.~\ref{fig:location_ablation} illustrates the visualization of words learned by each variant.
\begin{figure}[t]
\centering
\includegraphics[width=1\linewidth]{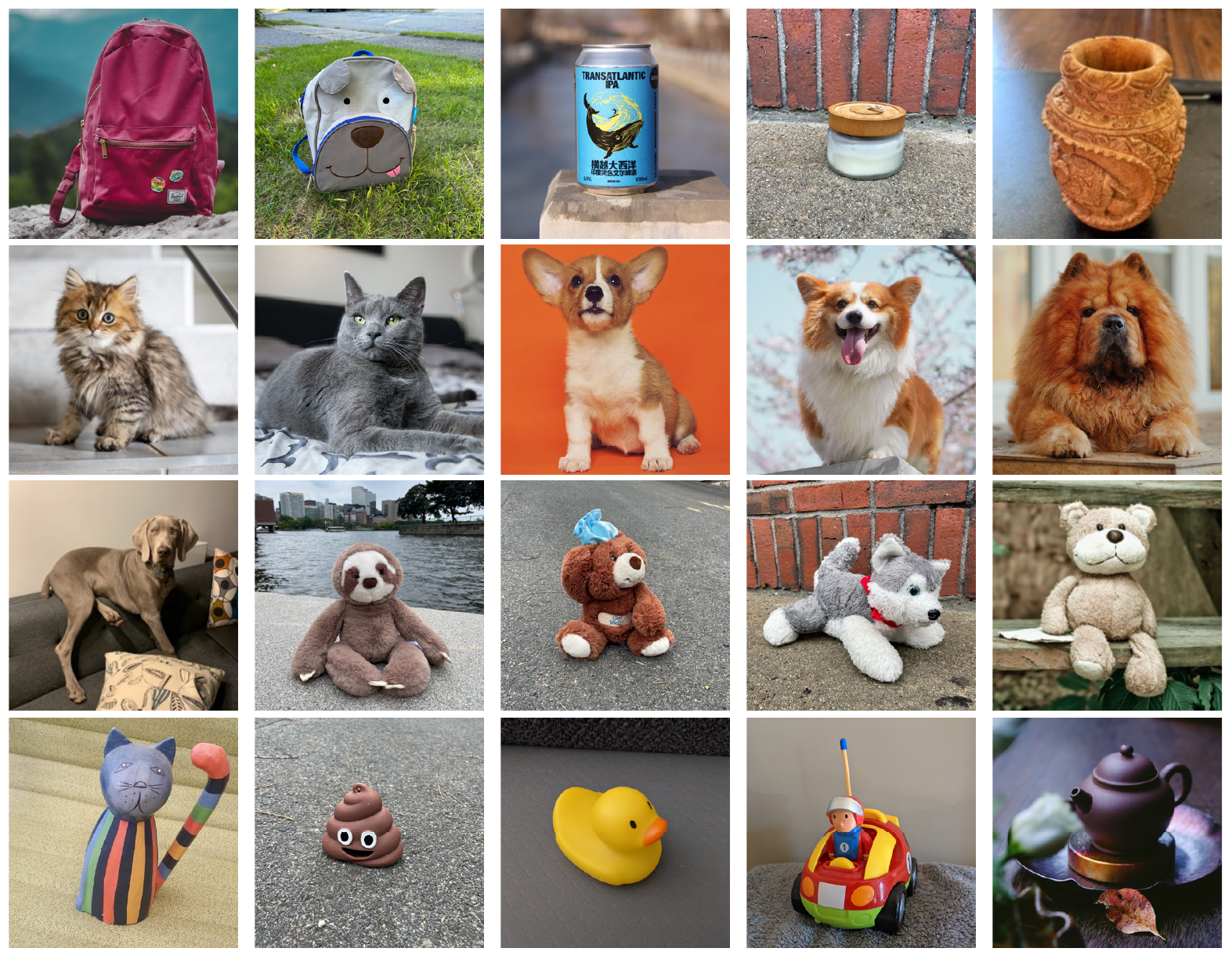}
\caption{Testing image samples.}
\label{fig:dataset}
\end{figure}
As shown in the figure, each variant contains one primary word that describes the subject concept.
When learning \emph{multiple word embeddings from multi-layer features}, we observe that the primary word is naturally linked to the features from the deepest layer, regardless of the position indices of layers.
Besides, as illustrated in Fig.~\ref{fig:location_ablation_editing}, in contrast to the primary word obtained by the single layer feature, the primary word learned by multi-layer features is well-editable.
Among them, our setting achieves better editability, which is shown in Table~\ref{tab:location_ablation}.

\begin{figure*}[t]
\centering
\includegraphics[width=1\linewidth]{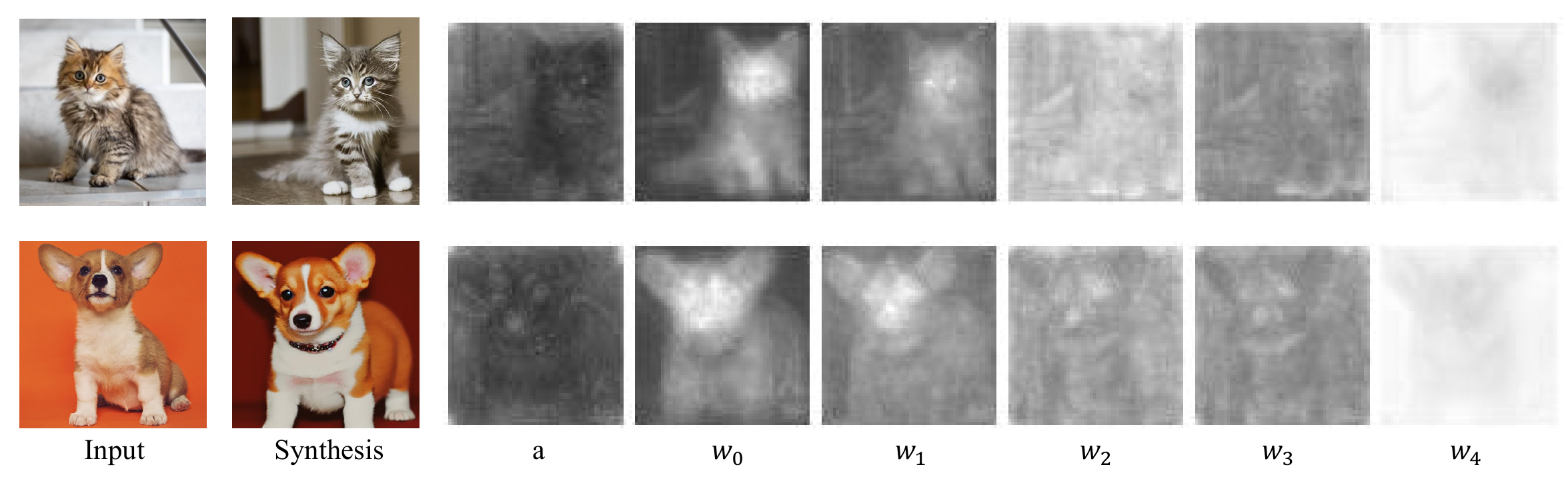}
\caption{Cross-attention map visualization. We show the average attention across timestep and layers for each word embedding. The attention map corresponding to the learned primary word $w_0$ delineates the subject region. }
\label{fig:attmap}
\vspace{-1em}
\end{figure*}

\begin{table}[t]
\begin{center}
\caption{\textbf{Ablation study}. [v] denotes the testing results are generated with full word embeddings $v$, while [w] denotes testing results are generated with the primary word embedding $w$. First, Middle, and Last denote the position of word embedding with respect to the deepest feature.}
\label{tab:location_ablation}
\resizebox{1\linewidth}{!}{
    \setlength{\tabcolsep}{2mm}
    \begin{tabular}{lccc}
        \toprule
        Method & CLIP-T ($\uparrow$) & CLIP-I ($\uparrow$) & DINO-I ($\uparrow$) \\
        \midrule
        Single-Layer Multi-Words [v] & 0.198 & 0.683 & 0.431  \\
        Single-Layer Multi-Words [w]  & 0.212 & 0.692 & 0.443 \\
        Multi-Layers Multi-Words Middle [v]  & 0.211 & 0.726 & 0.592  \\
        Multi-Layer Multi-Words Middle [w]  & 0.249 & 0.673 & 0.444  \\
        Multi-Layers Multi-Words Last [v]  & 0.217 & 0.722 & 0.585  \\
        Multi-Layers Multi-Words Last [w]  & 0.247 & 0.694 & 0.474  \\
        Multi-Layers Multi-Words First [v]  & 0.204 & \textbf{0.771} & \textbf{0.658}  \\
        Multi-Layer Multi-Words First [w]  & \textbf{0.257} & 0.699 & 0.486  \\
        \bottomrule
    \end{tabular}
}
\end{center}
\end{table}

\begin{figure*}[t]
\centering
\includegraphics[width=0.8\linewidth]{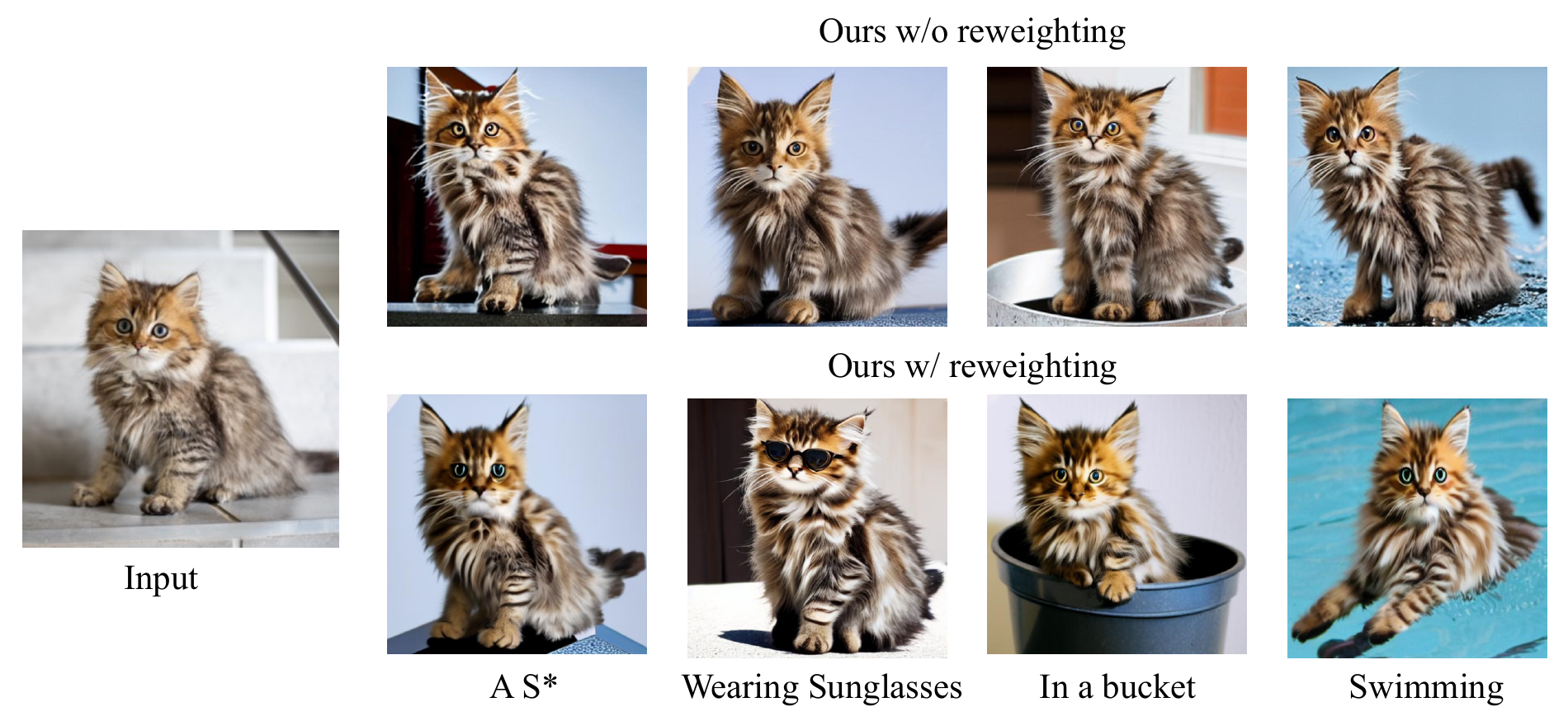}
\caption{Ablation of local attention map reweighting. Without the local attention reweighting, the learned model tends to be less editable.}
\label{fig:reweight}
\end{figure*}

\subsection{Effect of the local attention map reweighting}

The local mapping network aims to inject the fine-grained details of given subject during generation.
To further emphasize its effect on the subject region rather than irrelevant areas (\eg, background), we reweight the obtained local attention map by multiplying it with the attention map of primary word (refer to Sec.~{\color{red} 3.3} in main paper),
\begin{equation}
    A^l = A^l * \frac{A^g_{w_0}}{\max(A^g_{w_0})},
\end{equation}
where $A^l=\text{Softmax}\left(\frac{Q{K^l}^T}{\sqrt{d'}} \right)$ denotes the attention map of local mapping network and $A^g=\text{Softmax}\left(\frac{Q{K^g}^T}{\sqrt{d'}} \right)$ denotes the attention map of global mapping network.
$d'$ is the output dimension of key and query features.
$A^g_{w_0}$ is the attention map of primary word $w_0$.
To verify its effectiveness, we firstly visualize the cross-attention map of each word in input text prompt in Fig.~\ref{fig:attmap}.
As one can see, the learned primary word $w_0$ is associated with the subject concept and its attention map $A^g_{w_0}$ accurately delineates the subject region, so we can leverage it to reweight the local attention map $A^l$.
Furthermore, as illustrated in Fig.~\ref{fig:reweight}, without the local attention reweighting, the features of local mapping network may affect the subject-irrelevant areas, resulting in degraded editability.
In contrast, our ELITE with reweighting strategy reduces the disturbances on subject-irrelevant areas, and achieves better editability.

\begin{figure*}[t]
\centering
\includegraphics[width=0.9\linewidth]{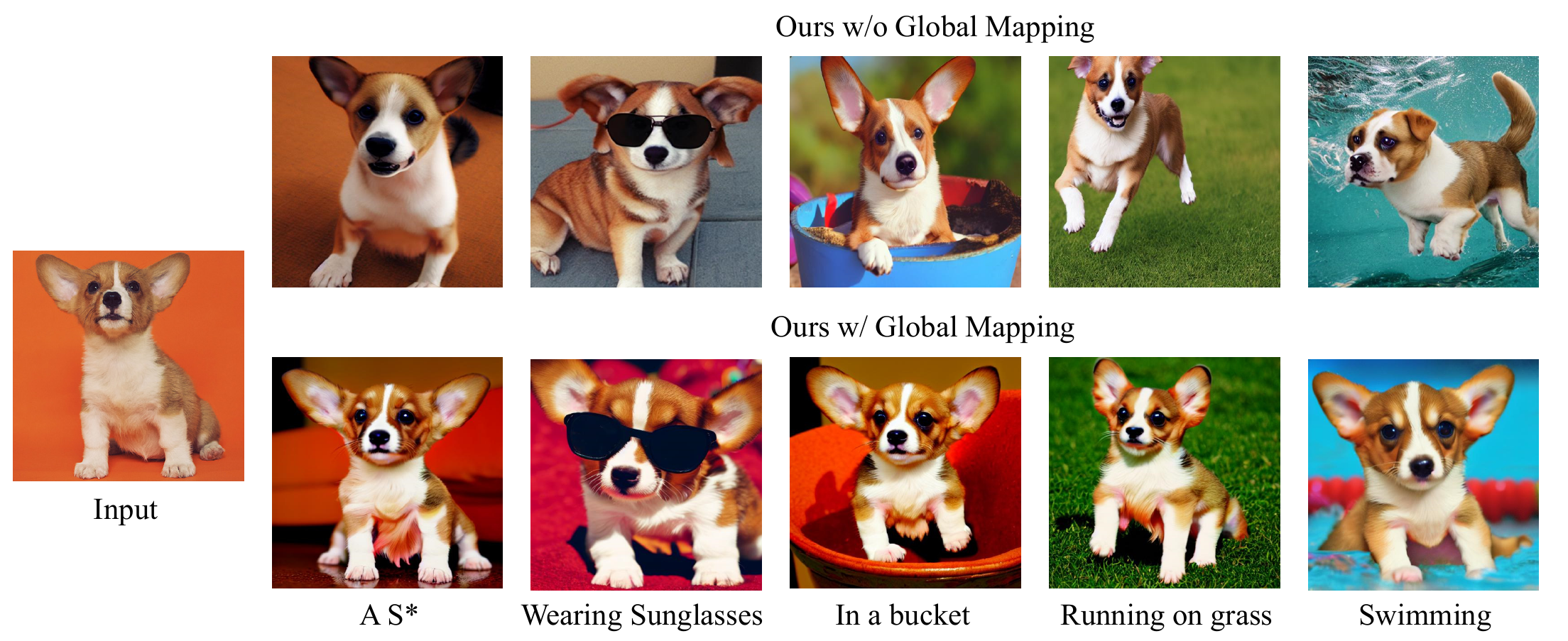}
\caption{Ablation of global mapping network. With the global mapping network, our ELITE faithfully recovers the target concept with higher visual fidelity while enabling robust editing.}
\label{fig:only_local}
\end{figure*}

\subsection{Effect of the Global Mapping}
We have further conducted the ablation study to evaluate the effect of our global mapping network.
For comparison, we remove the global mapping network, while replace the pseudo word \texttt{S*} with a ground-truth category word to learn the local mapping network (\eg, \texttt{S*} $\rightarrow$ \texttt{dog} in Fig.~\ref{fig:only_local}).
As shown in Fig.~\ref{fig:only_local}, without the global mapping, using the local mapping network only provides a few fine-grained details (\eg, fur color), yet fails to keep the structure of the concept (\eg, ear).
In contrast, by adding global mapping network to encode a suitable primary word embedding, our ELITE faithfully recovers the target concept with higher visual fidelity while enabling robust editing.

\section{More Experimental Details}
\label{sec:more_exp}

\subsection{Training Details}

\noindent \textbf{Textual Inversion~\cite{gal2022image}.} 
We use the official stable diffusion version of Textual Inversion\footnote{https://github.com/rinongal/textual\_inversion}.
For each subject, experiment is conducted with the batch size of 1 and a learning rate of 0.005 for 5,000 steps. 
The new token is initialized with the category word, \eg, ``cat''. 

\noindent \textbf{Custom Diffusion~\cite{kumari2022multi}.}
We use the official implementation of Custom Diffusion\footnote{https://github.com/adobe-research/custom-diffusion}. 
We train it with a batch size of 1 for 300 training steps.
The learning rate is set as 1e-5.
The regularization images are generated with 50 steps of the DDIM sampler with the text prompt ``\texttt{A photo of a [category]}''.

\noindent \textbf{DreamBooth~\cite{ruiz2022dreambooth}.}
We use the third-party implementation of DreamBooth\footnote{https://github.com/XavierXiao/Dreambooth-Stable-Diffusion}.
Training is done with finetuning both the U-net diffusion model and the text transformer. 
The training batch size is 1 and the learning rate is set as 1e-6. 
The regularization images are generated with 50 steps of the DDIM sampler with the text prompt ``\texttt{A photo of a [category]}''.
For each subject, we train it for 800 steps.

\subsection{Testing Datasets}

For customized generation, we adopt concept images from existing works~\cite{gal2022image, ruiz2022dreambooth, kumari2022multi} with 20 subjects, including dog, cat, and toy, \etc. 
Fig.~\ref{fig:dataset} illustrates the full image samples.

\subsection{Text prompts}

We adopt the text prompt list used in Textual Inversion~\cite{gal2022image} for training, which is provided as below:
\begin{itemize}
    \setlength{\itemsep}{0pt}
    \setlength{\parsep}{0pt}
    \setlength{\parskip}{0pt}
    \item ``a photo of a \pholder",
    \item ``a rendering of a \pholder",
    \item ``a cropped photo of the \pholder",
    \item ``the photo of a \pholder",
    \item ``a photo of a clean \pholder",
    \item ``a photo of a dirty \pholder",
    \item ``a dark photo of the \pholder",
    \item ``a photo of my \pholder",
    \item ``a photo of the cool \pholder",
    \item ``a close-up photo of a \pholder",
    \item ``a bright photo of the \pholder",
    \item ``a cropped photo of a \pholder",
    \item ``a photo of the \pholder",
    \item ``a good photo of the \pholder",
    \item ``a photo of one \pholder",
    \item ``a close-up photo of the \pholder",
    \item ``a rendition of the \pholder",
    \item ``a photo of the clean \pholder",
    \item ``a rendition of a \pholder",
    \item ``a photo of a nice \pholder",
    \item ``a good photo of a \pholder",
    \item ``a photo of the nice \pholder",
    \item ``a photo of the small \pholder",
    \item ``a photo of the weird \pholder",
    \item ``a photo of the large \pholder",
    \item ``a photo of a cool \pholder",
    \item ``a photo of a small \pholder",
\end{itemize}

For qualitative evaluation, we employ the editing templates used in~\cite{gal2022image, ruiz2022dreambooth, kumari2022multi}.
For quantitative evaluation, we employ the editing prompts from DreamBench~\cite{ruiz2022dreambooth}, which contains 25 editing prompts for each subject. 
The full prompts are listed in Table~\ref{tab:text-prompt}.

\begin{table*}[t]
  \caption{Text prompt list for quantitative evaluation.}
  \label{tab:text-prompt}
  \centering
  \begin{tabular}{ll}
    \toprule
    Text prompts for non-live objects & Text prompts for live objects \\
    \midrule
    ``a \pholder in the jungle'' & ``a \pholder in the jungle''\\
    ``a \pholder in the snow'' & ``a \pholder in the snow'' \\
    ``a \pholder on the beach'' & ``a \pholder on the beach'' \\
    ``a \pholder on a cobblestone street'' & ``a \pholder on a cobblestone street''\\
    ``a \pholder on top of pink fabric'' & ``a \pholder on top of pink fabric'' \\
    ``a \pholder on top of a wooden floor'' & ``a \pholder on top of a wooden floor'' \\
    ``a \pholder with a city in the background'' & ``a \pholder with a city in the background''\\
    ``a \pholder with a mountain in the background'' & ``a \pholder with a mountain in the background'' \\
    ``a \pholder with a blue house in the background'' & ``a \pholder with a blue house in the background''\\
    ``a \pholder on top of a purple rug in a forest'' & ``a \pholder on top of a purple rug in a forest''\\
    ``a \pholder with a wheat field in the background'' & ``a \pholder wearing a red hat'' \\
    ``a \pholder with a tree and autumn leaves in the background'' & ``a \pholder wearing a santa hat'' \\
    ``a \pholder with the Eiffel Tower in the background'' & ``a \pholder wearing a rainbow scarf'' \\
    ``a \pholder floating on top of water'' & ``a \pholder wearing a black top hat and a monocle'' \\
    ``a \pholder floating in an ocean of milk'' & ``a \pholder in a chef outfit'' \\
    ``a \pholder on top of green grass with sunflowers around it'' & ``a \pholder in a firefighter outfit''\\
    ``a \pholder on top of a mirror'' & ``a \pholder in a police outfit'' \\
    ``a \pholder on top of the sidewalk in a crowded street'' & ``a \pholder wearing pink glasses'' \\
    ``a \pholder on top of a dirt road'' & ``a \pholder wearing a yellow shirt''\\
    ``a \pholder on top of a white rug'' & ``a \pholder in a purple wizard outfit''\\
    ``a red \pholder'' & ``a red \pholder'' \\
    ``a purple \pholder'' & ``a purple \pholder'' \\
    ``a shiny \pholder'' & ``a shiny \pholder'' \\
    ``a wet \pholder'' & ``a wet \pholder''\\
    ``a cube shaped \pholder'' & ``a cube shaped \pholder''\\
    \bottomrule
  \end{tabular}
\end{table*}

\end{document}